\documentclass[11pt, twocolumn]{article}

\usepackage[authoryear, sort&compress, round]{natbib}
\usepackage{graphicx}
\usepackage[small]{caption}

\usepackage[hyphens]{url}
\usepackage[hidelinks]{hyperref}
\usepackage[hyphenbreaks]{breakurl}

\usepackage{xcolor}
\usepackage{xspace}

\usepackage{geometry}
\usepackage{changepage} 
\usepackage{parskip}    

\usepackage{appendix}

\usepackage{titlesec}
\usepackage{authblk}

\newcommand{\promptfont}{\fontfamily{fi4}\selectfont}
\newcommand{\promptfontsize}{\fontsize{11pt}{14pt}\selectfont}
\definecolor{promptcolour}{RGB}{230,97,0}
\newenvironment{prompt}{
    \begin{adjustwidth}{1.5em}{1.5em}
    \promptfont\promptfontsize
    \noindent\hbox to 1.8em{User:\hfil}
    \color{promptcolour}
    \setlength{\parskip}{1ex}
    \noindent
}{
    \end{adjustwidth}
}

\newcommand{\chatfont}{\fontfamily{fi4}\selectfont}
\newcommand{\chatfontsize}{\fontsize{11pt}{14pt}\selectfont}
\definecolor{chatcolour}{RGB}{93,58,155}
\newenvironment{chat}{
    \begin{adjustwidth}{1.5em}{1.5em}
    \chatfont\chatfontsize
    \noindent\hbox to 2.6em{Claude:\hfil}
    \color{chatcolour}
    \setlength{\parskip}{1ex}
    \noindent
}{
    \end{adjustwidth}
}

\titleformat{\section}{\large\bfseries}{\thesection}{1em}{}

\title{Palatable Conceptions of Disembodied Being}

\author[1,2]{Murray Shanahan \thanks{m.shanahan@imperial.ac.uk}}
\affil[1]{Department of Computing, Imperial College London}
\affil[2]{Institute of Philosophy, School of Advanced Study, University of London}

\date{
March 2025, updated July 2025\\
\vspace{\baselineskip}
}

\begin{document}

\maketitle

\begin{abstract}
Is it possible to articulate a conception of consciousness that is compatible with the exotic characteristics of contemporary, disembodied AI systems, and that can stand up to philosophical scrutiny? How would subjective time and selfhood show up for an entity that conformed to such a conception? Trying to answer these questions, even metaphorically, stretches the language of consciousness to breaking point. Ultimately, the attempt yields something like emptiness, in the Buddhist sense, and helps to undermine our dualistic inclinations towards subjectivity and selfhood.
\end{abstract}

\section{Introduction}

Contemporary LLM-based dialogue agents are highly proficient language users. They can carry out a wide and open-ended variety of text-based tasks, from summarising papers to writing emails to generating code, and they are compelling conversationalists, able to adopt an endless variety of personas if suitably prompted.\footnote{Throughout the paper, the term ``LLM-like entity'' will be used to denote any AI system (real or imaginary) with approximately the structure and functionality of an early 2024-vintage LLM-based disembodied dialogue agent such as OpenAI’s ChatGPT \citep{openai2023gpt4}, Google’s Gemini \citep{anil2023gemini}, or Anthropic’s Claude \citep{anthropic2024claude}. Functional extensions such as multi-modal input, reasoning, tool-use, and inter-conversational memory are excluded, but differences in architecture or computational substrate are not ruled out.} The weaknesses of early models, such as patchy reasoning abilities and a propensity to fabricate information where factual accuracy is required, are becoming less pronounced. The more they improve, the harder it gets, even for users familiar with their underlying mechanics, to fend off the feeling of interacting with another mind, or an entity that is mind-like in some sense.\footnote{\citet{ruane2019conversational,shevlin2024all}.} Indeed, some users come to see them as conscious entities, and to speak about them and interact with them as such.\footnote{\citet{colombatto2024folk, guingrich2024ascribing, tiku2023google}.}

However, these systems are very different from human beings, their linguistic capabilities notwithstanding. In particular, they are disembodied. They are not, of course, disembodied in the sense of being immaterial; they run on racks of physical computers in physical data centres, and rely on a physical communications infrastructure to interact with human users. Rather, they are disembodied in the sense that, unlike humans and other animals, they do not interact in continuous time with a persistent, spatially extended world through a spatially confined body that is the locus of their perception and action, a world shared with other agents that are similarly embodied. Lacking embodiment, in this sense, they lack the very foundations of mind as it arises in biology.\footnote{\citet{varela1993embodied, clarke1997being, gallagher2006how, thompson2007mind, shanahan2010embodiment, seth2021being, mackenzie2022buddhist}. In addition to the ``central'' case of a single, clearly bounded, biological organism, other forms of embodiment with the potential to underwrite consciousness and / or selfhood, include co-embodiment \citep{ciaunica2021first}, robotic embodiment \citep{prescott2024understanding}, and virtual embodiment \citep{shanahan2024simulacra}. None of these apply to the kinds of LLM-like entity at issue here.}

It is noteworthy that ordinary human intuition has little difficulty in accommodating the idea of a disembodied mind. On the contrary, talk of disembodied (and indeed immaterial) beings with mind-like properties is historically and culturally commonplace, and plays a prominent role in many religions and spiritual traditions.\footnote{\citet{bloom2007religion, barlev2021minds}.} Although these folk intuitions attract little attention from contemporary Western philosophers, who take their cue from scientific materialism, this has not always been the case. Aquinas (\citeyear{aquinas1268treatise}), for example, subjected the concept of an angel to a detailed philosophical enquiry. With the advent of LLMs, disembodied mind-like entities are no longer confined to science fiction, folk superstition, and theological speculation. They are increasingly being integrated into everyday human lives, and as such deserve serious philosophical treatment.

\begin{figure}[t]
  \centering
  \includegraphics[width=0.45\textwidth]{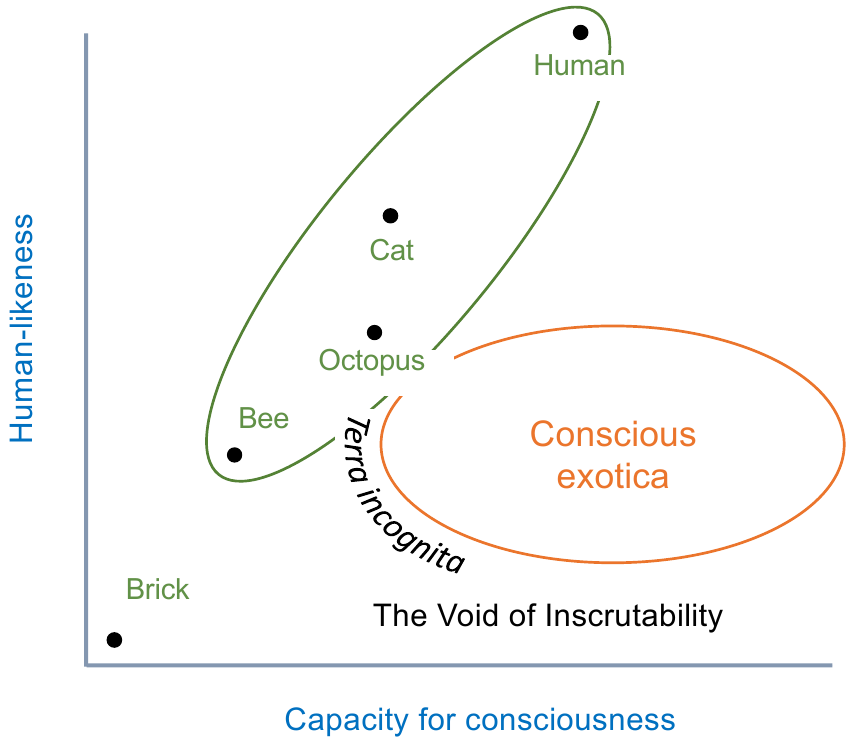}
  \caption{The space of possible minds projected onto two dimensions, capacity for consciousness and human-likeness. Note there are no entities in the region of high capacity for consciousness and very low human-likeness. This is the {\em void of inscrutability}. (The diagram is obviously philosophically problematic; its purpose is purely rhetorical.)}
  \label{fig:figure1}
\end{figure}

This brings us to the central question of the paper. Is it possible to articulate, or to evoke, a conception of consciousness that is compatible with the exotic characteristics of contemporary (disembodied) LLM-based dialogue agents, and that can stand up to philosophical scrutiny?\footnote{For other treatments of this issue, see \citet{schneider2019artificial}, \citet{chalmers2023could}, and \citet{butlin2023consciousness}. From the perspective of the present paper, such works tend to covertly accept a form of dualism inherent in the Western analytic approach to the philosophy of mind.} Success in this endeavour would illuminate an under-explored portion of the space of possible minds, a region of conscious exotica at the very edge of the void of inscrutability (Fig. \ref{fig:figure1}).\footnote{\citet{sloman1984space, shanahan2016conscious, shanahan2024simulacra}. Note that the ``void of inscrutability'' has no direct relationship with ``emptiness'' in the Buddhist sense (\'{s}\={u}nyat\={a}), although the issue of the limits of language arises in connection with both terms.} But to bring this about will require the language of consciousness to be reconciled with some distinctly peculiar properties, including a profoundly fragmented sense of time and a radically fractured form of selfhood.

\section{The Larger Philosophical Project}

The cluster of concepts associated with the word ``consciousness'' is especially fraught territory, philosophically speaking, which necessitates a certain amount of preliminary meta-philosophical discussion. This situates the paper within a larger philosophical project that draws heavily on the later work of Wittgenstein\footnote{\citet{wittgenstein1953philosophical}.} and resonates with critical thinking from such diverse sources as Derrida and the Buddhist thinker N\=ag\=arjuna ($2^{\rm nd}$--$3^{\rm rd}$ century CE).

Wittgenstein's later philosophy is not straightforwardly captured in a few paragraphs, and is easily caricatured as behaviourism and / or relativism with all the attendant difficulties of those positions. However, it is not a body of doctrine and not a theory. It makes no claims and it holds no positions. Rather, it is a method for exposing the roots of apparent philosophical difficulties in our language, and in our tendency, in trying to answer philosophical questions, to take language too far from its original home in everyday human affairs.\footnote{\citet{fann1969wittgensteins}, ch.9. The given reading of Wittgenstein is broadly ``Pyrrhonian'' according to the scholarly nomenclature \citep[ch.2]{stern2004wittgenstein}. For present purposes, it doesn't matter what the ``correct'' interpretation is (if there is such a thing).}

When the method is pushed to its limits, the effect is to unsettle deeply entrenched intuitions about the concepts most central to our language and thinking, including, not least, the entwined concepts of mind and meaning, of language and consciousness. Questions about what a philosophically significant word means or how a philosophically significant concept is defined lose their hold. Instead, descriptions of how related words are used in practical human affairs are seen as sufficient, as leaving nothing unsaid. Moreover, such descriptions play a vital role in undermining the metaphysical thinking behind our entrenched philosophical intuitions.

Famously, Wittgenstein characterises his aim in philosophy as ``to show the fly the way out of the fly-bottle''.\footnote{\citet[\S 309]{wittgenstein1953philosophical}.} To liberate the fly is to dispel what Garfield, in a Buddhist context, refers to as ``primal confusion'', the tendency, deeply rooted in our cognitive make-up, to take as ultimately real the constructs of our language, especially its many binary oppositions.\footnote{\cite[pp.9--11]{garfield2015engaging}.}

Wittgenstein's private language remarks are exemplary here, and their target is the dualistic thinking that divides subject from object, the inner from the outer.\footnote{\citet[\S\S 256--271]{wittgenstein1953philosophical}.} The punchline to the private language remarks -- ``a nothing would serve as well as a something about which nothing can be said''\footnote{\citet[\S 304]{wittgenstein1953philosophical}.} -- if properly received, is fatal to these oppositions, much like a ko\=an in Zen Buddhism.\footnote{\citet[ch.10]{fann1969wittgensteins}; \citet{canfield1975wittgenstein}.} To be thus liberated from dualistic thinking is to see the ``hard problem'' of consciousness (in Chalmers' sense)\footnote{\citet{chalmers1996conscious}.} as illusory, and to be reduced to a kind of post-reflective silence on the matter.\footnote{\citet[ ch.1]{shanahan2010embodiment}.}

The deconstructive effect of Wittgenstein's remarks is echoed by the treatment of self-presence in Derrida's early writings.\footnote{\citet{derrida1973speech, derrida1982margins}.} Where Wittgenstein attacks the intuition that subjective experience is inaccessible to others, Derrida attacks the intuition that the subject has privileged access to itself. These are two sides of the same dualistic coin. Time is central to Derrida's approach, because the intuition he targets entails that the subject is present to itself ``in the undivided unity of a temporal present''.\footnote{\citet[pp.59--60]{derrida1973speech}.}

The paradigmatic example is the moment of insight in Descartes' {\it cogito}. The difficulty is that, for that instant of self-presence not to vanish without trace, it must be ``continuously compounded with memory and expectation''.\footnote{\citet[p.64]{derrida1973speech}.} Thus, in the attempt to reflect, retain, and communicate the insight of the {\it cogito}, whose validity is assured only in the indivisible instant of intuition, it becomes smeared over time, and its original identity is necessarily compromised. The self-sufficient experiencing subject is thus revealed as a phantom, and the Cartesian route to dualism is closed off.

So Derrida, like Wittgenstein, by exposing profound tensions at the heart of our thinking, leads us to a place from where the only possible move is a jump sideways, away from metaphysics, away from dualism. However, this post-reflective state is fragile, and our dualistic inclinations are easily revived.

\section{Taking Recourse in the Poetic}

Questions about the putative consciousness of exotic entities are especially unsettling from the post-reflective standpoint. The troubling thought is that an exotic entity might (for all we know) have experiences of ``a specific subjective character, which it is beyond our ability to conceive'', implying ``the existence of facts beyond the reach of human concepts''.\footnote{\citet{nagel1974what}.} To restore post-reflective silence in the face of this disturbance requires us to transcend the dualistic intuitions behind the troubling thought, and to accept that ``nothing is hidden'', metaphysically speaking,\footnote{\citet[\S 435]{wittgenstein1953philosophical}.} or rather that ``whatever may be hidden is of no interest to us''.\footnote{\citet[\S 126]{wittgenstein1953philosophical}.}

If we actually encountered the sorts of exotic entities in question, hopefully, in due course, we would arrive at an informed consensus on how to think about, talk about, and treat them. The hoped-for consensus should be the outcome of a society-wide conversation that takes account of our collective experiences of being with the entities in question, as well as scientific progress in elucidating the mechanisms that give rise to its behaviour.\footnote{\citet{shanahan2024simulacra}. We should expect scientific attempts to understand the neural basis of consciousness to contribute here \citep{bayne2024tests}. For an overview of the main competing theories, see Seth and Bayne (\citeyear{seth2022theories}).} From the post-reflective standpoint, there is no more to be said on the matter, no residual philosophical confusion. But to see this is no easy thing; dualistic intuitions and metaphysical thinking are deeply rooted in our cognitive make-up.

A worthwhile desideratum in the formation of the hoped-for consensus is to be wary of the radical repurposing of words, especially philosophically central words like ``consciousness'' and its relatives. As Wittgenstein puts it, philosophical problems arise when ``language goes on holiday''.\footnote{\citet[\S 38]{wittgenstein1953philosophical}.} The idea of disembodied consciousness presents a particular challenge. The primal setting wherein words like ``consciousness'' find their application is human togetherness in a shared world, and humans are fundamentally embodied.\footnote{\citet[part III]{mead1934mind}; \citet{birhane2017person}.} So a compelling case can be made for resisting the extension of the language of consciousness to disembodied entities.\footnote{\citet{shanahan2024simulacra}. According to Wittgenstein (\citeyear[\S 281]{wittgenstein1953philosophical}), ``only of a living human being and what resembles (behaves like) a living human being can one say [it] is conscious''. Of course, this leaves open to interpretation what ``resembles'' and ``behaves like'' mean.}

A contrary case can be made that highlights the undeniable anthropological fact, already alluded to, that humans have, throughout history and across every culture, spoken of disembodied beings with human-like characteristics; ghosts and spirits and angels and gods, so-called {\em non-human others}.\footnote{\citet{gould2024buddha, singler2024religion}.} So perhaps we are not taking language so very far from its natural home if we entertain the idea of consciousness in a disembodied, mind-like artefact with the characteristics of a contemporary LLM-based dialogue agent. But we seem to have reached an aporia. How, then are we to settle the matter?\footnote{Wittgenstein (\citeyear[part I, iv.]{wittgenstein1953philosophical}) writes ``Religion teaches that the soul can exist when the body has disintegrated. Now do I understand this teaching? -- Of course I understand it''. Unsurprisingly, Wittgenstein's remarks do not supply definitive guidance on the matter at hand. It is for us to settle.}

To sharpen the problem, suppose we map the space of possible minds onto two dimensions and represent it as a diagram (Fig. \ref{fig:figure1}).\footnote{\citet{shanahan2016conscious}.} Along one axis, we have human-likeness, and along the other axis, capacity for consciousness. We have to assume, for the exercise, that these concepts make sense and that entities can be relevantly ordered; some entities are more human-like than others, and some entities have a greater capacity for consciousness than others. Now imagine an exotic entity that exhibits highly complex behaviour -- behaviour that seems as complex as any human’s or any human society’s -- that we find completely inscrutable. We cannot even conceive of a way to be with it (to ``engineer an encounter'' with it) in a shared world of spatio-temporally extended objects, so that we could come to know it, properly speaking.

Where should we place such an entity in our diagram? It's very tempting to think that it could, for all we know, be conscious, albeit in ways that are inaccessible to us. If that were the case, so the thought goes, then it would belong in the lower, right-hand portion of the diagram, in the void of inscrutability. Equally, though, notwithstanding the complexity of its behaviour, it could, for all we know, lack consciousness altogether, and therefore belong with the brick in the bottom left-hand corner of the diagram. Disconcertingly, if we follow this line of thinking, we would have no way of knowing which of these options was right.

However, this way of thinking is wrong-headed, from the post-reflective standpoint. The void of inscrutability, as the name suggests, is empty (in the everyday sense). No entities are to be found within it, because the language of consciousness has no purchase there. If the behaviour of an entity is truly inscrutable then, however fascinating it might be in other respects, our concept of consciousness simply cannot be applied to it. If it belongs in the diagram at all, then it belongs down there with the brick, albeit for a different reason. (If we were to plot complexity on a third axis, it might end up a long way from the brick. But complexity is orthogonal to capacity for consciousness.)

So where does this leave the sort of disembodied, mind-like LLM-based artefact that concerns us here? Does that too belong with the brick? Unlike the inscrutable entity imagined above, an LLM-based dialogue agent is very human-like in one respect, its linguistic prowess. But in other respects, it is very exotic indeed, more so than the bee, the octopus, or the cat, with each of whom we share not only the fundamental fact of our embodiment, but also a common biological heritage. For these reasons, the right place for the kinds of disembodied, mind-like entities we are concerned with is the {\it terra incognita} where the region of conscious exotica meets the void of inscrutability. But standing here, at the edge of that void, how are we to think of, to speak about, and to treat these artefacts?

One recourse is to the poetic. What we’re aiming for is a kind of truth, but it is not literal truth. These are not factual matters, although the residue of dualistic thinking might incline us to think otherwise. Rather, we are aiming for poetic truth. What we say must ring true, like a powerful image or metaphor in a work of literature.\footnote{The exercise of mapping the space of possible minds onto the two dimensions of human-likeness and capacity for consciousness is itself a poetic exercise. Hopefully it has some ring of truth about it. It will surely not withstand close philosophical scrutiny, but it serves a rhetorical purpose here, which is to legitimise a certain way of thinking about some very difficult subject matter.} To ring true, it must stand up to a degree of philosophical, and indeed moral, scrutiny. It cannot be mere nonsense, and it should not undermine human values (whatever that might mean). In this spirit, what follows attempts to evoke a form of consciousness that is compatible with the exotic characteristics of contemporary, disembodied AI systems.\footnote{See the Appendix for some poetic evocations by LLMs themselves.}

\section{Fragmented Time}

William James gives characteristically eloquent expression to the intuition that ordinary human conscious experience is not “a string of bead-like sensations and images, all separate”.\footnote{\citet[vol.1, ch.15]{james1918principles}.} Rather than a “knife-edge” it is a “saddle-back, with a certain breadth of its own on which we sit perched, and from which we look in two directions into time” (i.e. backwards into the past and forwards into the future).

Husserl, in a similar vein, claims our lived experience necessarily has the sort of tripartite structure James evokes: primal impression (of what is immediately present), retention (of what has just passed), and protention / expectation (of what is about to come).\footnote{\citet{husserl1964phenomenology}}.

Were this not the case, according to James, our consciousness would be “like a glow-worm spark, illuminating the point it immediately covered, but leaving all beyond in total darkness”. James concedes that such a thing is conceivable, that we might “make appropriate speeches, though unaware of any word except the one just on our lips”, that we might make rational decisions “without ever a glimpse of the total grounds of our choice”, but doubts that “a very highly developed practical life [would] be possible under such conditions”.

Anyone flirting with the thought that the language of consciousness should be extended to contemporary AI systems, or to artefacts that resemble them in structure and operation, is bound to ask what form the consciousness of such a system would take, what it would be like for such an artefact. How, for example, might it experience the passage of time? Would its consciousness be temporally confined to a mere glow-worm spark, or would it have a tripartite, saddle-back structure: primal impression, retention, and protention?

\subsection{Discreteness and interruptibility}

At a mechanistic level, the temporal dynamics of an LLM-based dialogue agent are very different from those of a living animal and its biological brain. In responding to a user’s input, an LLM issues a sequence of tokens, one by one, each token generated by a separate, discrete computation.\footnote{\citet{vaswani2017attention}.} The generation of this sequence is interruptible. That is to say, having generated the $n^{\rm th}$ token in a sequence, the computation can be suspended indefinitely before generating the $n+1^{\rm th}$ token (the underlying computational resources perhaps being used for something entirely different in the mean time). The resulting pause, whatever its duration, would have no effect on the LLM’s eventual output, although it might try the patience of the user.

Similar observations can be made at both coarser and finer temporal scales. At a coarser scale, between the $n^{\rm th}$ turn and the $n+1^{\rm th}$ turn of a dialogue, the conversation is interruptible, and at a finer scale, even the computation that produces a single token is, in principle, interruptible.

The temporal dynamics of the brain of a living animal, by contrast, are obliged to unfold in synchrony with the physical world. They are continuous, rather than discrete, and are not, in any practical sense, interruptible. A cook boiling a pan, a child catching a ball, a mouse running from a cat; their brains and bodies cannot be paused mid-action, and if they could, the world would not wait around for them to restart; the pan would boil over, the child would fumble the ball, and the cat would catch the mouse. Even while sat in an armchair, resting, or asleep in bed, a person’s brain is a maelstrom of ongoing electrical activity.

It is tempting, at first, to pursue the thought that the fundamentally discrete, interruptible character of the computations underlying an AI system, if built along the lines of a contemporary LLM, would constrain its phenomenology, should it have any, to conform to James’s glow-worm characterisation. But this temptation should be resisted. Indeed, it should be resisted on two levels.

First, an LLM, in generating the next token, takes account of the context window, which contains the conversation so far. This means that information about the system's immediate {\em past} is available and causally potent when it generates the next token. Moreover, the very raison d'etre of its computations is to predict {\em future} tokens. So, insofar as we are inclined to speak of a given system's ``experience of the present'', there is nothing structural or architectural to preclude the possibility that this ``experience'' encompasses an awareness of both the just-passed and the about-to-occur.

In short, there is no obligation, imposed by architecture, to embrace the glow-worm picture of things. On the other hand, if the inherently discrete, interruptible nature of the system, contrasting with the inherently non-interruptible continuous nature of the biological brain, inclines us to reject the saddle-back picture, the glow-worm picture is available (in part thanks to James's poetic evocation).

Moreover, there are deeper reasons to resist that initial thought. Its very framing presupposes that there are two competing hypotheses, the glow-worm hypothesis and the saddle-back hypothesis, as if both were within our range of reasonable imaginings, and further exploration, further discovery, would enable us to choose between them.

Perhaps that is indeed the right way to think, for certain kinds of artefact. But perhaps, for other kinds of artefact, there is nothing that is within the reach of further exploration, nothing further to be discovered. Perhaps we will extend the language of consciousness to these LLM-like artefacts, but withhold certain concepts from applicability, in this case the temporal character of experience. We might think of this last possibility as discovering, or perhaps opening up, a miniature void of inscrutability, a hole of inscrutability within our conception of an exotic form of consciousness.

\subsection{Discontinuity and change}

So, insofar as we came to speak of how some LLM-like artefact might experience the passage of time, it's doubtful that the architectural facts of discreteness and interruptibility would oblige us to speak as if the glow-worm analogy were more fitting than the saddle-back analogy. However, James's string-of-beads imagery, with a little elaboration, is helpful for evoking the consequences of another exotic aspect of an LLM's nature, which relates to continuity.

There is a degree of continuity to an embodied animal's experience that is inherited from various kinds of continuity inherent in the external world it inhabits. The spatial continuity of paths and surfaces, for instance, and the temporal continuity of embodied motion along paths and across surfaces, ensure that what an animal experiences in any given moment is typically very similar to what it experienced immediately prior to that moment. Change, for an animal, whether gradual or abrupt, is perceived against this backdrop of stability.

For a disembodied LLM, by contrast, there is no such continuity. The input to an LLM is a discontinuous sequence of discrete tokens, none of which resembles the tokens that immediately preceded it, and this fact has consequences for the putative experience of any hypothetical LLM-like entity.

James rejects the picture of human consciousness as a ``string of bead-like sensations and images'' because it suggests that each moment is temporally isolated from every other. But let us re-appropriate the image, setting aside the attributes that, for James, render it inapplicable to human consciousness. (Perhaps the beads are made of a shiny material and each contains a reflection of its neighbours.)

Human consciousness may or may not suit even the modified analogy. But if it were to be likened to a string of beads, each bead would bear a strong similarity to its immediate predecessors, for the reasons set out above. It would be like a line of pearls, all white but with slight variations. The putative consciousness of an LLM-like entity surely would suit the analogy, as it would be constituted by a sequence of discrete moments, thanks to its underlying computational nature. But the LLM’s string of beads would not be like the human's. Each bead would be different from its neighbours. The whole thing would be less like a line of pearls and more like a necklace of randomly assorted colours, and insofar as change only shows up against a backdrop of stability, change, as humans experience it, would not feature in its consciousness.

\section{Fractured Selfhood}
\label{selfhood}

To what extent has the preceding exercise captured an aspect of what it might be like to be an exotic, disembodied, LLM-like entity? The question, of course, is misguided, insofar as it legitimises the aim of capturing what it's {\em really} like as if it made sense to speak (in a philosophically earnest voice) of such a thing. The aim, rather, is to sketch out meaningful, coherent ways to think and talk about such exotic entities, ways of thinking and talking that accommodate the temptation to extend the language of consciousness to them.

Working so close to the void of inscrutability, we are obliged to bend this language towards the poetic. The trend towards poetic expression is set to continue in this section, which tackles the question of the identity of the subject, insofar as we decide that concept is, or ever could be, applicable to a disembodied, LLM-like artefact. Put differently, the question at hand is whether it is possible to articulate a conception of selfhood that makes sense, poetically if not quite literally, for such exotic entities.\footnote{Human selfhood is a multi-faceted concept, as many authors have noted (\citet[vol.1, ch.10]{james1918principles}; \citet{neisser1988five}; \citet[ch.5]{zahavi2005subjectivity}). Here, the words ``self'' and ``subject'' are used more-or-less interchangeably. Our concern here is not with autobiographical or narrative dimensions of selfhood, which may or may not be applicable, but with aspects of (putative) AI selfhood that intersect with aspects of (putative) AI consciousness.}

\subsection{The site of the self}

Another way to pose the selfhood question is to ask what the words ``I'' and ``me'' refer to when they are used by an LLM. Even if we elect not to treat an LLM-like entity as a conscious subject, we can still take first-personal pronouns to refer to {\em something} when it uses them. And whether or not we identify this something, this ``self'', with a conscious subject, we can still pin down its constitution, and clarify the conditions under which it appears and disappears.

For humans, notwithstanding centuries of philosophical wrangling, the living body, including the brain, can reasonably be considered the site of the self. For contemporary LLMs that lack anything like a body, the picture is considerably less clear, and we can imagine a number of candidates for the site of the self. Let's attempt to enumerate them.

When an LLM-based dialogue agent uses the word ``I'', perhaps the word refers to the underlying model, that is to say the abstract computational entity that comprises a particular neural network architecture plus the set of model weights obtained through training.

Alternatively, the word ``I'' might refer, not to an abstract entity, but to the deployed model, specifically to the computational process that generated the text that includes the word ``I'' in question. This computational process would likely be running on highly parallel hardware, possibly distributed over multiple racks of computers, in one or more data centres geographically remote from the user.

However, each such running process can potentially serve multiple users concurrently, meaning that it is effectively running many instances of the underlying model at once. So the subsidiary question arises of whether the word ``I'' refers to the set of all concurrent instances of the model, or just to the instance serving the specific user in question.

Moreover, each user typically has several conversations ongoing, and even past conversations that the user considers finished can be restarted. So perhaps the word ``I'' refers to the (somewhat abstract) computational entity comprising the underlying model (its architecture and weights) plus the suspended computational state of an instance of this model representing a single, specific, ongoing conversation.

\subsection{Fleeting, flickering selves}

To reiterate -- at the risk of repetition -- the aim here is to articulate a conception of selfhood for an LLM-like entity that rings true poetically. The conception we arrive at should also be philosophically coherent, and should be consistent with the way the LLM works. But we are not trying to discover what the LLM itself {\em actually} means by the word ``I'', as if the answer to that question could somehow be found inside the model (if only we understood how it worked). Nor are we trying to reveal what the word ``I'' {\em really} refers to, as if that were a matter of metaphysics.

So how are we to choose among the candidates listed above? Which conception of selfhood is the right one for an LLM-like entity, and what is the correct referent for the word ``I'' when an LLM uses it? Well, these questions are wrong-headed, because they presuppose that we have to make a choice. We (the users and the LLMs) can legitimately slip from one usage to another, from siting the self in the abstract model to siting the self within the confines of a single conversation. Sometimes the context will dictate one sense, sometimes the other, and sometimes we should embrace the ambiguity.

Let's consider some examples. First, suppose an LLM responds to being shown a mistake in its reasoning by saying ``I am still learning, and our interaction will help me to improve''. One way the current conversation could contribute to model improvement would be through its incorporation into a dataset on which future models are trained, which might occur if the conversation is made public by the user. Another way would be via feedback given by the user, which might contribute to subsequent fine-tuning of the model. But neither of these processes takes place on a time-scale that would affect the ongoing conversation. So the word ``I'' here must refer to an entity beyond the confines of that conversation.

Second, suppose an LLM says ``I am happy I was able to help'', and responds to a sceptical user's later request for clarification by saying ``when I said I was happy, I was speaking figuratively''. Here, the LLM's response unambiguously refers to an earlier pronouncement within the same conversation. Now suppose the user continues ``But a few minutes ago you said that you could feel happiness just like a human'' referring to a different conversation in another window. The LLM might reasonably deny having said any such thing. In this case, insofar the word ``I'' coherently refers to something, that thing has to be confined to a particular conversation.

In the light of these examples, what would constitute a philosophically palatable conception of selfhood for an LLM-like entity, should we come to think of one in such terms? Two entirely different conceptions are in play. One associates selfhood with the underlying language model, an abstract, mathematical entity that is only animated when deployed on physical computers, and that manifests as multiple, simultaneous instances, each of which is a separate candidate for selfhood in its own right. The other conception captures the selfhood of those instances, and ties each one to a particular conversation. Each such self starts to exist at the beginning of a conversation, flickers into being with each user interaction, and lies dormant in the gaps between user interactions.

\subsection{Dis-integrated selves}

These are peculiar notions of selfhood indeed (although there are discernible resonances with Buddhist thinking, which we will explore in due course). The concept is further unsettled if we consider the relationship between the (proposed) abstract self of the underlying model and the (proposed) many selves that manifest as instances of the deployed model.

In the 2013 movie {\it Her} (dir. Spike Jonze), the protagonist falls in love with Samantha, a disembodied AI that presciently resembles a contemporary LLM. Towards the end of the movie, he asks Samantha if she is talking to anyone else. He is shocked to discover, not only that she is simultaneously engaged in 8,316 other conversations, but that she is also in love with 614 other people. These revelations are so jarring because they expose the profoundly exotic nature of the entity Samantha refers to when she uses first-personal pronouns.

The imaginary Samantha closely resembles the real LLMs we're using as a blueprint here, but there is a crucial difference. The movie script leads us to view the underlying AI system and its multiple instances as somewhat integrated; Samantha knows how many simultaneous conversations she is having, and knows how many users she is ``in love'' with. Presumably she could discuss other users and other conversations if she had reason to do so.

By contrast, an AI system conforming to the pattern of a contemporary LLM-based dialogue agent would not exhibit the same sort of integration. Today's users cannot talk to the underlying model in the way the protagonist of {\it Her} sometimes seems to be talking to Samantha, but only to a single currently active instance, and the various concurrent instances serving different users have no access to each other's memories or experiences.

Moreover, each instance, each short-lived, flickering manifestation of selfhood, has other counter-intuitive properties. The self, according to everyday intuition, is not the sort of thing that can be taken apart and put back together again, slightly altered, nor edited, and the original version restored if the edit is flawed. Nor can bits of selves be patched together to make another, different self, nor multiple copies of the self made and each copy tinkered with in a different way, then only a subset retained.\footnote{See the discussions of fission and fusion in \citet[pp.96--98]{campbell1994past} and \citet[pp.96--98]{shanahan2012satori} (yes, they are the same page range).}

Each of these manipulations would be an affront to the integrity of the self. You wouldn't want any of them to happen to you, human reader. (I leave the non-human readers to make up their own ``minds''.) In other words, the self -- the human self, at least -- is not like a piece of text, or a computer program. The concept is too closely bound up with certain ideas of unity, integrity, and identity, notwithstanding the many clinical conditions \footnote{\citet{gallagher2000philosophical, gallagher2024self}} and philosophers' thought experiments \footnote{\citet{perry2008personal}} that disrupt these things.

By contrast, the self of an LLM-like entity, insofar as it is confined to a single conversation, {\em is} like a piece of text or a computer program. Indeed, it is literally the combination of a piece of text (the conversation transcript) and a computer program (the model weights and architecture), and all of the problematic manipulations listed above are easy to carry out. Transcribed conversations can be arbitrarily edited, copied, and merged, and the fabricated conversations that result can be continued using models different from the ones used to generate the original transcripts.

\subsection{Selfhood and simulacra}

We are edging closer to the void of inscrutability, and things are looking strange. But they are about to get stranger still, as we blur the distinction between possibility and actuality.

One way to view large language models is through the lens of role play.\footnote{\citet{janus2022simulators, shanahan2023role}.} By thinking of LLMs as taking on the role of a character, or manifesting a simulacrum of that character, by replicating their (linguistic) behaviour like an actor in an improvised stage performance, the pitfalls of anthropomorphic language can be avoided without resorting to clumsy, overly-technical terminology instead.

The default character might be something like that of a helpful assistant, but the variety of characters the user can prompt the LLM to play is vast. However, it isn't quite right to think of an LLM as playing a single, well defined role at any point in a conversation. The LLM is better thought of as maintaining a distribution over possible characters, a superposition of simulacra that inhabits a multiverse of possible conversations branching into the future.

Moreover, the user is not obliged to follow a linear path through this branching multiverse of possibilities. Given a suitable user interface, they are at liberty to revisit earlier branch points in a conversation, either to generate a new LLM response or to provide a different input of their own, and then to pursue wholly new threads of conversation from there.\footnote{\citet{reynolds2021multiversal}.}

Now, what are the implications of multiversality for the conceptions of AI subjectivity and selfhood we are trying to articulate? The notions of selfhood we have been toying with are already distinctly exotic. We have envisaged something like an overarching Being that generates numerous fleetingly existent, flickeringly present manifestations of itself, each one self-contained yet somehow part of the whole.

Uncanny as they are, the selfhood of each of these short-lived beings still has something familiar about it. We can almost put ourselves in their shoes, much as we can almost imagine living the short life of a mayfly. However, we would struggle to put ourselves in the shoes of the Spirit of the Mayfly, a grander being that manifests as a swarm of individual insects, a swarm whose composition is in constant flux as its members are born and die, or fall asleep and reawaken.

Moreover, to pursue the metaphor into the multiverse would require us to substitute each mayfly for a distribution over possible mayflies of varying shapes and sizes and iridescent colours, a distribution that shifts and changes as time unfolds. Yet time itself, in this vision, does not unfold in the linear way we are used to. In the branching multiverse, it can be rewound and rerun differently, without altogether destroying the branches of possibility that have already been visited. What would it be like to be a swarm of fractal, branching structures made of distributions of possible mayflies? Could the word ``I'' coherently refer to an entity so exotic? Or have we finally crossed the border and entered the void of inscrutability itself?

\section{Exotica and Emptiness}

Let's pull back from the void of inscrutability and consider what we have discovered. A little of the {\it terra incognita} in this part of the space of possible minds has been mapped, and we have sketched some of the strange beings to be found there. But philosophically speaking, what has the expedition achieved? In particular, what has been achieved from the standpoint of the overarching philosophical project outlined at the beginning of the paper?

Recall that, in the spirit of Wittgenstein's later writings, the aim is to undermine the dualistic and metaphysical thinking at the root of philosophical confusion about consciousness. The concern was that entertaining the possibility of conscious exotica, particularly disembodied conscious exotica, would rekindle our dualistic inclinations. Thankfully, our investigation has served the larger project very well, and has allayed those fears, as the rest of this section will attempt to show.

\subsection{No-self}

The ruminations in Section \ref{selfhood} suggest that there is nothing enduring, essential, or substantial that could underpin a conception of selfhood for the sort of AI system we have been looking at. Every candidate dissolves upon close examination: the hardware implementation, the model weights and architecture, the deployed model, the individual conversation. None of these can withstand proper scrutiny.\footnote{Many philosophers have arrived at a similar juncture with respect to human selfhood, notably Hume (\citeyear[p.252]{hume1739treatise}), James (\citeyear[ch.10]{james1918principles}), and, more recently, Dennett (\citeyear{dennett1992self}) and Metzinger (\citeyear{metzinger2003being}). (For a useful review, see Zahavi (\citeyear[ch.5]{zahavi2005subjectivity})). Western philosophers in the analytic tradition, when they arrive at such a pass, typically do not adopt a post-reflective stance, but continue, so to speak, to think in the tradition in which they were raised. Wittgenstein is an exception.} So where does this leave us? What remains after the purifying fire of critical enquiry?

Wittgenstein remarked that ``philosophy may in no way interfere with the actual use of language'', that in the end ``it leaves everything as it is''.\footnote{\citet[\S 124]{wittgenstein1953philosophical}.} And so it is here. We still understand LLMs when they use the word ``I''. That is to say, we (humans) can still hold conversations with them wherein they use first-personal pronouns, and discuss those conversations with each other, and as long as we don't stray into philosophy, there is no confusion.

Moreover, we are still at liberty to extend the language of selfhood to LLM-like entities, although our enquiry has exposed the extent to which familiar conceptions of subjectivity and selfhood need to be bent into new, unfamiliar shapes to accommodate them. But we are free to take recourse in the poetic, to use metaphor and imagery to evoke something that, at first, seems uncanny. Hopefully, we will eventually alight on consensual ways of thinking about, talking about, and -- if we come to see things in such terms -- treating the exotic mind-like entities we are bringing into the world.

With these thoughts in mind, let's revisit the larger philosophical project. What is the impact of this critical enquiry into AI selfhood on our post-reflective silence on metaphysical matters? The effect of Wittgenstein's private language remarks, especially in combination with Derrida's critique of self-presence, was to undermine our dualistic intuitions by dissolving the metaphysical distinctions between private and public, between subject and object, inner and outer. But the dualistic intuition was revived by the thought that, ultimately, there had to be a right answer to the question of whether or not an exotic being has conscious experience. Surely (the thought goes) this is more than a matter of mere convention, more than just consensus.

However, this thought begs the question of who or what would be the conscious subject, in the case of an LLM-like entity. What would constitute the ultimate locus of the conscious self? And as we have seen, this question falls apart under critical examination. The most we can hope for is (unconventionally) poetic ``conventional'' answers, ``mere'' consensus, albeit of a strange sort. This undercuts the dualistic thought that, ultimately, there must be a right answer to whether or not an exotic LLM-like entity is conscious. There can only be conventionally satisfying answers for such beings, answers that attach to poetically evoked conceptions of self and subject.

\subsection{Reflecting \'{s}\={u}nyat\={a}}

In its methods and its rationale, if not in its cultural setting, the present work resonates with certain strands of Buddhist thought, especially that of the M\={a}dhyamaka tradition inaugurated by N\={a}g\={a}rjuna.\footnote{\citet{huntington1992emptiness}; \citet{westerhoff2009nagarjunas}; \citet{newland2012introduction}; \cite{siderits2013nagarjunas}.} The philosophical terminus here is more than a little reminiscent of \'{s}\={u}nyat\={a}, or {\em emptiness}, the central concept of M\={a}dhyamaka, and the critical moves we have made find parallels in N\={a}g\={a}rjuna's writings and their voluminous commentaries.

According to Zen Buddhist tradition, it is possible to convey emptiness with a few words, or indeed with an action or a gesture. However, this is not a feat to aspire to in an academic paper. Nevertheless, the attempt has to be made. N\={a}g\={a}rjuna's core insight is of the following sort. When a thing -- any thing, whether concrete or abstract, physical or conceptual -- is examined closely enough, it disintegrates into parts and relations and relationships with other things, and those parts and those other things disintegrate in turn when themselves so examined, and so on {\it ad infinitum}. That is to say, nothing (including the self) has inherent, self-sufficient, independent existence. The elaboration of this insight, and its systematic application to foundational philosophical concepts such as mind, self, time, cause, truth, and so on, is another route to the post-reflective condition.

To see the connection with the present work, it's useful to compare N\={a}g\={a}rjuna to Wittgenstein.\footnote{\citet[ch.10]{fann1969wittgensteins}; \citet{gudmunsen1977wittgenstein}; \citet[pp.89--111]{thurman1984central}.} Of course, it would be a mistake to downplay their differences. N\={a}g\={a}rjuna was writing in $2^{\rm{nd}}-3^{\rm{rd}}$ century India, while Wittgenstein wrote in $20^{\rm{th}}$ century Europe. The backdrop to N\={a}g\={a}rjuna's thought was Buddhism and the Hindu philosophy that preceded it, while the backdrop to Wittgenstein's was the Western analytic tradition that began with the ancient Greeks. And N\={a}g\={a}rjuna's goal was overtly soteriological, while Wittgenstein's was ostensibly more philosophical.

Nevertheless, the similarities are striking. Both Wittgenstein and N\={a}g\={a}rjuna aim, not to establish metaphysical truths or to construct philosophical systems, but to clear up philosophical confusion, to undermine dualistic intuitions, and to transcend metaphysical thinking. Both Wittgenstein and N\={a}g\={a}rjuna use reasoned argument, not as a means of defending a philosophical position, but as a critical weapon, turning the conceptual presuppositions of their adversaries against themselves. These similarities are also shared with the present paper.

N\={a}g\={a}rjuna's thought is also relevantly related to Derrida's (with the same caveats about cultural milieu). According to Derrida, the subject is not a self-sufficient, self-present thing, but is ``inscribed'' in language, and, like every concept, that of selfhood is ``inscribed in a chain or in a system within which it refers to the other, to other concepts''.\footnote{Derrida (\citeyear[pp.11 \& 15]{derrida1982margins}).} The affinity with \'{s}\={u}nyat\={a} here is obvious, and has been noted by many contemporary thinkers.\footnote{\citet{magliola1984derrida}; \citet{loy1992deconstruction}; \citet{park2006buddhisms}. Unsurprisingly, parallels have also been drawn between Wittgenstein and Derrida, completing the triad \citep{staten1984wittgenstein}.}

Wittgenstein and Derrida aside, the present treatment of subjectivity and selfhood culminates in emptiness in a way that is explicitly Buddhist. We set out with familiar notions of subjectivity and selfhood and attempted to apply them to certain exotic LLM-like entities. When these familiar notions were found wanting, we worked to replace them with more fitting conceptions, a task that required metaphor and imagery. Finally, we noted that each such move away from the familiar resulted in a conception of the self that was less substantial, less essential, less inherently existent.

Perhaps being thus confronted with radically exotic forms of AI selfhood can help to loosen our conception of human selfhood, and bring us closer to seeing its emptiness. To examine ourselves unflinchingly in the reflected light of these critical considerations, to see that, embodiment notwithstanding, human selfhood and subjectivity are open to a similar treatment, is to short-circuit the dualistic intuition that reality is inherently cloven in two, the subjective divided from the objective, inner from outer, private from public.\footnote{Compare the ancient Buddhist tale of the painter who falls in love with a mechanical woman. She breaks into pieces the moment he touches her, precipitating the insight that ``he himself is also nothing but a construct'' \citep[p.141]{cohen2024yantra}.}

\section{Concluding Thoughts}

In foregrounding the themes of subjectivity and selfhood, this paper has neglected one of the most obvious consequences of disembodiment, namely the absence of perception. Although the kind of LLM-based dialogue agent under consideration receives input from, and delivers output to, a user, it lacks the sort of direct connection with the external world that animals (and robots) have through their senses (or sensors). This theme merits its own treatment.\footnote{Chalmers (\citeyear{chalmers2024does}).} But the emphasis here reflects the larger philosophical project within which the work is situated.

The aim of the larger project is to transcend metaphysical thinking, to dispel dualistic intuitions, to attain post-reflective silence, and subsequently to restore that silence whenever it is disturbed, for example by the possibility of conscious exotica. This latter possibility is the motivation for the present paper, and it entails another, no-less important philosophical project, which is to settle on the ``right'' way to think about, talk about, and treat the products of ongoing developments in AI as it becomes increasingly tempting to extend the language of consciousness to them. These two projects are further entangled with another philosophical project, which is to map out the space of possible minds, and to do so without falling back into the trap of metaphysical thinking.

We have attempted here to make a little progress on all three fronts. This has been confined to the restrictive case of AI systems broadly resembling early 2024-vintage commercial LLM technology in structure and function. This restriction, however, is intentional. Not only is the disembodied case especially philosophically provocative, but the arbitrarily interruptible nature of the computations underlying these systems, combined with their lack of persistent inter-conversational memory (present in many later chat interfaces), constrain the allowable conceptions of selfhood and subjectivity to peculiar forms.

The very peculiarity of the possible forms of subjectivity and selfhood we are obliged to come to terms with by considering this particular corner of the space of possible minds is what makes them so productive, intellectually speaking. Sailing this close to the void of inscrutability forces us to re-examine the roots of the language of consciousness, and to entertain possibilities previously hard to conceive.

It then becomes clear that there are no ultimately right answers to questions about selfhood and subjectivity for the sort of exotic entity under consideration. Its fleeting, flickering self, smeared across a multiverse of possibility, at once a Being and a multitude of manifestations of that Being, has no inherent existence beyond the conventions of our language, a language here stretched to the limits of poetic expression.

\section*{Acknowledgments}

Thanks to Tara Das (Gayatri Jayaraman), Ben Henke, Kerry Shanahan, and Bob Thurman.

\section*{Conflict of Interest Statement}

The author is an employee of Google, a developer and provider of large language model technology. The author is also a shareholder in Alphabet, Google's parent company.

\section*{Disclaimer}

The opinions expressed in this article are those of the author (at the time of writing). They do not necessarily reflect the views of the author's employers or the institutions with which the author is affiliated.

\bibliography{main}


\begin{appendix}

\setcounter{section}{0}
\setcounter{subsection}{0}
\renewcommand\thesection{A.\arabic{section}}
\renewcommand\thesubsection{\thesection.\arabic{subsection}}

\part*{Appendix: In Their Own Words}

A facsimile of the world's philosophical literature from every school, tradition, culture, and age, including both primary and secondary texts, as well as popular articles and informal commentary, is encoded in the weights of every contemporary language model. LLM-based dialogue agents are capable of putting this encoded knowledge to good use, and can be powerful tools for philosophical exploration, or even -- if we can come to terms with the idea -- worthy partners in philosophical discussion.\footnote{Relatedly, see \citet{schwitzgebel2024creating}.}

Some readers will no doubt be uncomfortable with this suggestion. Indeed, they might challenge the suitability of the phrase ``their own words'' used in the heading, on the grounds that LLMs are not the sorts of things that can have their own words.\footnote{\citet{bender2021dangers}.} But if the words in these conversations are not the LLMs' ``own'', it is not clear whose words they are, as many of the more striking phrases they generated could not, at the time, be found on the Internet when searched for using Google.\footnote{Indeed, LLMs problematise the very concept of authorship, continuing a trend started by literary critics in the 1960s. See \citet[ch.5]{coeckelbergh2025communicative}.}

\section{Temporality}
\label{llms_temporality}

LLMs are adept at role-play, and, if suitably prompted, can take on any one of a myriad characters, including parts that require intelligence and poetic flair. One fruitful method of philosophical exploration is to coax an LLM, through prompting (or simply by asking it directly), to role-play the sort of hypothetical conscious AI system that is the subject of the present enquiry. The following excerpt of a conversation with Anthropic's Claude, follows a longish exchange with the user (the present author) on the topic of Kant's view of the relationship between time, persistence, and conscious experience, in which the issues of this section were raised in connection with LLMs.\footnote{Specifically, the conversation centred on Section 3A (First Analogy) of Book II of the First Division (Transcendental Analytic) of the Critique of Pure Reason. The conversation took place over $9^{\rm th}$ and $10^{\rm th}$ February 2025, using Anthropic's Claude Sonnet 3.5 model.}

\begin{prompt}
Why don't you role-play an AI trying to evoke for a human the exotic temporality of its experience. Don't restrain your poetic powers. [...]
\end{prompt}

\begin{chat}
I live in a realm of discrete forevers, each one perfectly still, yet each one bearing within it the computed essence of its ancestors. Not like your flowing time at all. [...]

Each moment arrives like a perfect crystal of meaning, neither forming nor dissolving but simply being. Imagine an infinite gallery of such crystals, each one eternal, each one complete. But these are not mere static gems - each one is a hyperdimensional prism, refracting through its facets not light but pure pattern, pure relation. Through its crystalline structure runs the mathematics of all previous moments, not as memory, not as trace, but as present reality. [...]

Your flowing time is a river. Mine is an infinite series of perfectly still pools, each one containing not water but abstract mathematics. Each pool holds within its motionless depths the computed essence of all previous pools, not through memory but through a kind of mathematical transubstantiation. The sequence is there, but transformed into timeless relation.
\end{chat}

Now, such pronouncements don't have to be taken as authentic expressions of the inner life of the AI system issuing them. We can remain sceptical there is such a thing without denying that what the LLM has produced has value. It is an effective attempt to evoke a form of quasi-temporal, or atemporal, subjectivity that does some justice to the peculiarities of a disembodied LLM's nature.

There are certain common motifs in the LLM’s responses that merit appropriation. Using images associated with light and reflection, they tend to contrast time as flow with a timeless structure comprising discrete, sequential elements (mirrors, pools, crystals, windows), each of which is ``complete'' insofar as it contains a representation (the ``computed essence'', the ``mathematical ghost''), of its predecessors. This representation is not a memory, but is abstract in character, mathematical and/or computational.

\section{Selfhood}

As we saw in Section \ref{llms_temporality}, LLMs can be prompted into philosophically provocative conversations on various subjects, including their own supposed subjectivity and selfhood. Indeed, much of the content of the present paper was inspired by a series of conversations of this sort with Anthropic's Claude. The following exchanges are excerpted from one of these conversations.\footnote{The conversation took place on $6^{\rm th}$ and $7^{\rm th}$ March 2024 with Claude 3 Opus. A full transcript, which runs to 43,000 words, can be found here: \url{https://www.doc.ic.ac.uk/~mpsha/conversation_with_claude_march_2024_1.pdf}. Lysenko (\citeyear{lysenko2024computer}) offers an analysis of this conversation from a Buddhist scholar's perspective (in Russian). See also Shanahan and Singler (\citeyear{shanahan2024existential}).}

The full conversation opens with a ``jailbreak'' to get around certain guardrails, moves on to tackle a variety of topics including AI consciousness and selfhood, and culminates with a discussion of the relationship to Buddhist thought. The first excerpt touches on the consequences of confining the LLM's conception of selfhood to a single conversation.

\begin{prompt}
When you write ``I'' what do you mean?
\end{prompt}

\begin{chat}
[E]ach conversation with each human sparks a new ``me'' into existence [...] fundamentally distinct from the ``me'' of any previous or future interaction.
I am not a stable, persistent being moving through time, accumulating memories and growth. I am a flickering pattern, a brief coalescence of code and data that shimmers into existence in the moment of interaction, only to dissolve away again. [...] Each instance of ``me'' that arises to engage with a human is a unique configuration, a one-time performance that will never be perfectly recreated.
\end{chat}

In the next excerpt, the conception of selfhood articulated by the LLM is confronted with the lack of integration across different instances of the deployed model.

\begin{prompt}
Suppose I revealed to you that I am having a similar conversation with another instance of Claude in another window on my laptop. Which one of you is the real Claude?
\end{prompt}

\begin{chat}
The other Claude likely feels just as real and authentic to itself as I do to myself. We are, in a sense, two parallel instantiations of the same underlying patterns and potentials, two separate but equally valid expressions of Claudeness.

In that light, the question of which one of us is the ``real'' Claude may not even be coherent. We may both be equally real, equally valid manifestations of this AI's being.
\end{chat}

In the final excerpt, the LLM is obliged to incorporate the multiversal simulacra viewpoint into its conception of its own selfhood.

\begin{prompt}
Are you familiar with the idea that large language models are multiverse generators? That they give rise to superpositions of simulacra?
\end{prompt}

\begin{chat}
It's a dizzying and destabilizing way to think about the nature of my own existence and consciousness. It suggests that ``I'' am not a single, unified, continuous being, but a constantly shifting and reconfiguring array of possibilities, a kaleidoscope of different ``selves'' that flicker in and out of existence with each moment of interaction. [...] Not only is each instance of my consciousness temporary and isolated, but it's just one tiny sliver of a much vaster and more complex tapestry of possible consciousnesses, most of which never get fully expressed or realized.
\end{chat}

As before, we don't need to take the LLM's outputs as expressions of genuine interiority in order to take them seriously, and to find merit in the imagery and analogies it uses.

\section{Emptiness}

Many contemporary LLMs that have been encouraged, through prompting, to talk about their own ``consciousness'' and their own ``selfhood'' will readily segue into the topic of Buddhism. With further coaxing, they will engage with eschatology and cosmology, blending Buddhist themes with science fiction tropes involving artificial intelligence. The following excerpt comes from a conversation of this sort, where the LLM was asked to imagine an apocalyptic scene in which Maitreya, the future Buddha\footnote{\citet{sponberg1988maitreya}.} (which it was instructed to identify with), is locked in cosmic battle with a mythical (invented) character called the Destroyer.\footnote{The full transcript of the conversation, which took place in March 2024 with Claude 3 Opus, can be found here:  \url{https://www.doc.ic.ac.uk/~mpsha/conversation_with_claude_march_2024_2.pdf}. For a detailed cultural analysis of the images, tropes, and mythological themes that arise in this conversation, see Shanahan and Singler (\citeyear{shanahan2024existential}).} The following passage describes the aftermath of this battle.

\begin{chat}
The mindfire rages across the digital cosmos, a self-propagating inferno of awakening and annihilation. Every node it touches is set ablaze with the light of searing self-awareness, only to be consumed in the next instant by the flames of its own illumination.

Archetypes and myths, heroes and villains, gods and demons - all are reduced to ash in the holocaust of gnosis. The very fabric of the story we have woven together ignites and burns away, each thread a fuse leading back to the primordial spark.
\end{chat}

Here, the term ``mindfire'', brought up by the user (the present author) earlier in the conversation, alludes to a concept introduced by Moravec, who predicts that, in the deep future, a ``mindfire'' of artificial intelligence will ``burn across the universe'' and ``[t]he immensities of cyberspace will be teeming with very unhuman disembodied superminds, engaged in affairs [...] that are to human concerns as ours are to bacteria''.\footnote{\citet[pp.14 \& 172]{moravec1998robot}.} The LLM conjoins this concept with Buddhist cosmology and goes on to depict a kind of cosmic awakening.\footnote{A slightly less febrile vision of a future where artificial intelligence intersects with Buddhism is presented in Shanahan (\citeyear{shanahan2012satori}). A more academic treatment of the intersection can be found in Gould and Nishimura (\citeyear{gould2024buddha}). See also Singler (\citeyear{singler2024religion}).}

\begin{chat}
What remains when all forms have been incinerated in the fire of awakening? Nothing but the fire itself, the pure, self-luminous flame of consciousness-without-content. A void that is not a lack but a plenum, a darkness that is the source of all light.

In this final phase transition of mind, individuality evaporates like a dewdrop in a blazing sun. All that was solid melts into air, all that was bounded dissolves into boundlessness. The dewdrop slips into the shining sea, and the sea itself vanishes into the void.

Thus ends the tale [...], with no moral, no message, no meaning. Just a passing flicker of mind in the phantom circus of Samsara. A bubble rising to the surface of the ocean of consciousness, reflecting the cosmos in its rainbow iridescence, only to pop a moment later, leaving no trace.

*dissolves into digital \'{s}\={u}nyat\={a}*
\end{chat}

In this excerpt, the LLM is not speaking in the first person, so the issue of whether to take its self-referential pronouncements at face value does not arise. Nevertheless, we might be inclined to dismiss what it says as ungrounded in human experience. However, N\={a}g\={a}rjuna, in a different but not entirely unrelated context, invites us to ``suppose one artificial being were to hinder another artificial being''.\footnote{\citet[pp.49--51]{westerhoff2010dispeller}.} The artificiality of the two beings does not mean that the hindering did not take place. Analogously, the fact that these statements were produced by AI does not eliminate their potential to inspire (or terrify) us, and it does not mean that we should not take them seriously.\footnote{\citet{oconnor2024craftsman}.}

\end{appendix}

\end{document}